\documentclass[conference]{IEEEtran}
\usepackage{cite}
\usepackage{amsmath,amssymb,amsfonts}
\usepackage{algorithmic}
\usepackage{graphicx}
\usepackage{textcomp}
\usepackage{xcolor}
\def\BibTeX{{\rm B\kern-.05em{\sc i\kern-.025em b}\kern-.08em
    T\kern-.1667em\lower.7ex\hbox{E}\kern-.125emX}}

\usepackage{multirow}
\usepackage{comment}
\usepackage{subfigure}

\newcommand{\archtab}{% Just for this example
\resizebox{0.55\columnwidth}{!}{%
\begin{tabular}{c|c|c|c}
\hline
\textbf{Network} & \textbf{Layer}   & \textbf{Parameters}     & \textbf{In/Out} \\ \hline
\textbf{$E_S$}      & Conv-1D          & k = 8, s = 2            & 45/45           \\
\textbf{}        & IC + LeakyReLU   & dropout = 0.05          & -               \\
\textbf{}        & Conv-1D          & k = 8, s = 2            & 45/25           \\
\textbf{}        & IC + LeakyReLU   & dropout = 0.05          & -               \\
\textbf{}        & Conv-1D          & k = 8, s = 2            & 25/15           \\
\textbf{}        & IC + LeakyReLU   & dropout = 0.05          & -               \\
\textbf{}        & Conv-1D          & k = 8, s = 2            & 15/10           \\
\textbf{}        & IC + LeakyReLU   & dropout = 0.05          & -               \\
\textbf{}        & Conv-1D          & k = 8, s = 2            & 10/4            \\
\textbf{}        & IC + LeakyReLU   & dropout = 0.05          & -               \\ \hline
\textbf{$E_A$}      & Conv-1D          & k = 8, s = 2            & 45/96           \\
\textbf{}        & IC + LeakyReLU   & dropout = 0.05          & -               \\
\textbf{}        & Conv-1D          & k = 8, s = 2            & 96/128          \\
\textbf{}        & IC + LeakyReLU   & dropout = 0.05          & -               \\
\textbf{}        & Conv-1D          & k = 8, s = 2            & 128/128         \\
\textbf{}        & IC + LeakyReLU   & dropout = 0.05          & -               \\
\textbf{}        & Conv-1D          & k = 8, s = 2            & 128/128         \\
\textbf{}        & IC + LeakyReLU   & dropout = 0.05          & -               \\
\textbf{}        & Conv-1D          & k = 8, s = 2            & 128/64          \\
\textbf{}        & IC + LeakyReLU   & dropout = 0.05          & -               \\ \hline
\textbf{$D$}       & UP + Conv1D      & scale = 2, k = 7, s = 1 & 68/128          \\
\textbf{}        & IC + LeakyReLU   & dropout = 0.05          & -               \\
\textbf{}        & UP + Conv1D      & scale = 2, k = 7, s = 1 & 128/128         \\
\textbf{}        & IC + LeakyReLU   & dropout = 0.05          & -               \\
\textbf{}        & UP + Conv1D      & scale = 2, k = 7, s = 1 & 128/128         \\
\textbf{}        & IC + LeakyReLU   & dropout = 0.05          & -               \\
\textbf{}        & UP + Conv1D      & scale = 2, k = 7, s = 1 & 128/96          \\
\textbf{}        & IC + LeakyReLU   & dropout = 0.05          & -               \\
\textbf{}        & UP + Conv1D      & scale = 2, k = 7, s = 1 & 96/45           \\
\textbf{}        & IC + LeakyReLU   & dropout = 0.05          & -               \\ \hline
\textbf{$C_A$}      & Linear + ReLU    & -                       & 16/32           \\
\textbf{}        & Dropout + Linear & dropout = 0.5           & 32/4            \\ \hline
\end{tabular}
}}
    
\begin{document}
\bstctlcite{BSTcontrol}

\title{Towards Explainable, Privacy-Preserved Human-Motion Affect Recognition\footnotemark{\textsuperscript{1}\\
} 

}

% {\footnotesize \textsuperscript{*}Note: Sub-titles are %not captured in Xplore and
% should not be used}
% \thanks{Identify applicable funding agency here. If none, %delete this.}
%}

\author{\IEEEauthorblockN{Matthew Malek-Podjaski}
\IEEEauthorblockA{\textit{School of Computing Science} \\
\textit{University of Glasgow}\\
Glasgow, United Kingdom \\
2323841m@student.gla.ac.uk}
\and
\IEEEauthorblockN{Fani Deligianni}
\IEEEauthorblockA{\textit{School of Computing Science} \\
\textit{University of Glasgow}\\
Glasgow, United Kingdom \\
fani.deligianni@glasgow.ac.uk}
% \and
% \IEEEauthorblockN{3\textsuperscript{rd} Given Name Surname}
% \IEEEauthorblockA{\textit{dept. name of organization (of Aff.)} \\
% \textit{name of organization (of Aff.)}\\
% City, Country \\
% email address or ORCID}
% \and
% \IEEEauthorblockN{4\textsuperscript{th} Given Name Surname}
% \IEEEauthorblockA{\textit{dept. name of organization (of Aff.)} \\
% \textit{name of organization (of Aff.)}\\
% City, Country \\
% email address or ORCID}
% \and
% \IEEEauthorblockN{5\textsuperscript{th} Given Name Surname}
% \IEEEauthorblockA{\textit{dept. name of organization (of Aff.)} \\
% \textit{name of organization (of Aff.)}\\
% City, Country \\
% email address or ORCID}
% \and
% \IEEEauthorblockN{6\textsuperscript{th} Given Name Surname}
% \IEEEauthorblockA{\textit{dept. name of organization (of Aff.)} \\
% \textit{name of organization (of Aff.)}\\
% City, Country \\
% email address or ORCID}
}

\maketitle
\addtocounter{footnote}{1}
\footnotetext
{Authors acknowledge funding from EPSRC EP/R045178/1}
\addtocounter{footnote}{1}
\footnotetext
{Codebase will be released upon acceptance.}

\begin{abstract}
Human motion characteristics are used to monitor the progression of neurological diseases and mood disorders. Since perceptions of emotions are also interleaved with body posture and movements, emotion recognition from human gait can be used to quantitatively monitor mood changes. Many existing solutions often use shallow machine learning models with raw positional data or manually extracted features to achieve this. However, gait is composed of many highly expressive characteristics that can be used to identify human subjects, and most solutions fail to address this, disregarding the subject's privacy. This work introduces a novel deep neural network architecture to disentangle human emotions and biometrics. In particular, we propose a cross-subject transfer learning technique for training a multi-encoder autoencoder deep neural network to learn disentangled latent representations of human motion features. By disentangling subject biometrics from the gait data, we show that the subject's privacy is preserved while the affect recognition performance outperforms traditional methods. Furthermore, we exploit Guided Grad-CAM to provide global explanations of the model's decision across gait cycles. We evaluate the effectiveness of our method to existing methods at recognizing emotions using both 3D temporal joint signals and manually extracted features. We also show that this data can easily be exploited to expose a subject's identity. Our method shows up to 7\% improvement and highlights the joints with the most significant influence across the average gait cycle.  
\end{abstract}


\section{Introduction}
\author{mikulincer_embodiment_2015}
Strong neuroscientific evidence \cite{mikulincer_embodiment_2015} shows that there is an interaction between brain networks involved in gait and emotion. It does not come as a surprise that certain gait characteristics are related to mood disorders, such as depression and anxiety. Gait analysis has revealed several indices that correlate to emotional well-being. For example, increased gait speed, step length and arm swing have been related to positive emotions. In contrast, a low gait initiation reaction time and flexion of posture have been linked to negative feelings \cite{hashmi_motion_2020,deligianni_emotions_2019}. 

However, developing robust algorithms for human motion and gait analysis requires monitoring patients continuously at their home. This poses tremendous ethical and privacy challenges that have to be addressed in order for the technology to be successfully accepted and adapted.
Privacy-preserving deep learning technologies for computer vision applications is a relatively new research area that is usually based on disentangling biometric features representation from other attributes \cite{chen_privacy_2021,gong_disentangled_2020}. 
Some old-fashion techniques of face obfuscation such as pixelation, blurring and masking to protect privacy offer limited protection while they compromise the ability to track human motion precisely. For example, inference in deep learning models is compromised if there is an uncertainty on whether the view is anterior or posterior with relation to the human's head. Furthermore, facial features also encode relevant information, such as the subject's mood.

To disentangle human biometrics and affects, we exploit a Multi-encoder Autoencoder structure that achieves cross-subject transfer learning as inspired by recent works by Gu et al. \cite{gu_cross-subject_2020}, who highlighted their efficiency for disentanglement learning in abnormal gait recognition based on joints movements of the lower limbs. This concept is also similar to motion retargeting, which allows transferring motions from one subject to another by disentangling motion, skeleton, and camera view data \cite{aberman_learning_2019}. To our knowledge, this is the first study that disentangles affects and biometrics and demonstrates that this approach preserves subjects privacy while it improves classification performance. 

Furthermore, the ability to provide insight into which factors are affecting the output and by how much is crucial for developing explainable machine learning models and a necessary condition in evaluating and accepting these models in healthcare applications \cite{arrieta_explainable_2019}. We choose a gradient-based explainability method, which is inherently model-specific and local, to obtain intuitive visualisation of the deep neural network. Subsequently, we aggregate information across testing samples in each class which allows us to inspect global explanations of the model.

Overall, the contribution of our work is threefold: a) we introduce a novel method to disentangle affects and biometrics from gait data, which significantly improves the performance of the affect recognition model, b) we demonstrate that our architecture of disentangling affect and biometrics enhances subject's privacy considerably and c) we exploit a gradient-based neural network visualisation method to examine global explanations of the model across classes. This reveals the joints that contribute significantly to the prediction and identify which phases of the gait cycle are most informative.

\section{Related Work}

\begin{figure}[htbp]
    \centering
    \includegraphics[width=.9\linewidth]{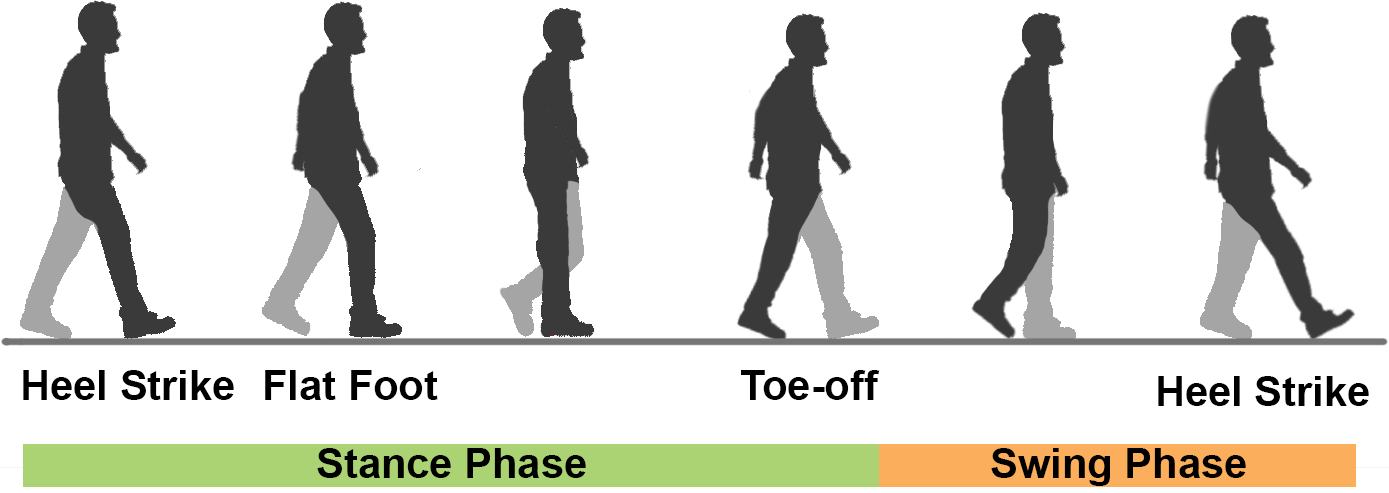}
    \caption{An illustration of a single gait cycle, highlighting key events of the right foot and the two phases (stance and swing) that compose the gait cycle.}
    \label{fig:gait_cycle}
    \vspace{-0.7em}
\end{figure}

\subsection{Gait Cycle}
Human gait involves repetitive sets of motions, iterating with each step. When analysing such motions we confine the descriptions to a single gait cycle, defined as the period between two consecutive heel strikes of the same heel \cite{vaughan_gaitcd_1999}. A single gait cycle is composed of two main phases, the stance phase, and the swing phase. With the stance phase being composed of the single limb stance, during which only one foot is touching the ground, and a double support where both feet are touching the ground \cite{xu_emotion_2020}. The stance phase begins as the heel strikes the ground and ends with the toe-off, when the foot lifts off the ground. Subsequently, the gait enters into the swing phase, the phase with the foot off the ground, which is broken down into the initial swing, mid swing and the terminal swing. The latter happens as the heel strikes the ground for the second time. A visual representation of the gait cycle along with the key gait events is shown in Figure \ref{fig:gait_cycle}.

\subsection{Emotion within Gait}
Embodiment theory as described by Winkielman et al. \cite{mikulincer_embodiment_2015}, "is the idea that higher-level processing is grounded in the organism's sensory and motor experiences", suggesting that the perception of emotions and cognitions is interleaved with body postures and movements. For example, when people recall joyful memories, they partially reproduce that state. Embodiment theory suggests that such re-enactments are crucial for identifying and interpreting the differences between affects. Most importantly, they distinguish that such re-enactments are not always conscious. This concept builds the foundation that human motions can be used to identify internal emotional states.

% In further support of this motion-emotion entanglement, other research has shown that not only does this relation exist but is, in fact, bi-directional. For example, the effects of body posture changes on a subject's emotional response have been a popular subject of study. Hackford et al. \cite{hackford_effects_2019} have presented a study showcasing this phenomenon where subjects showed improved psychological state as a result of walking with an upright posture—strengthening the idea that emotions and their physical re-enactments are deeply correlated. 

\subsection{Gait Emotion Characteristics}
Extensive research has been done into manually extracted features from human motion in order to understand how humans perceive emotion from body language.
Studies such as those by Gross et al. \cite{gross_effort-shape_2012}, show that when a subject's specific target emotions are stimulated during their gait, those same emotions are subconsciously re-enacted through the subject's motions, and could be identified reliably by human observers at a ratio greater than chance. The same study also identified an exhaustive list of quantitative gait metrics and their importance to each emotion. In particular, metrics such as: velocity, stride length, hip/shoulder/elbow flexion, and pelvis/trunk rotation were shown to have statistically significant differences between emotions. 
Similar studies \cite{roether_critical_2009, montepare_identification_1987, barliya_expression_2013} found stride lengths and gait speed to be a significant characteristic, as well as increased arm swing, thigh elevation angles and cadence being strongly correlated with happiness and anger.

\subsection{Gait Subject Biometrics}
Aside from neurological conditions, gait is also highly expressive regarding subjects' identity. In fact, subject's gait styles usually dominate over any subtle neurological effect. There is significant research on identifying subjects and their gender based on their gait characteristics \cite{loula_recognizing_2005}. 
% \cite{lu_human_2014, stevenage_visual_1999, loula_recognizing_2005, shiqi_yu_study_2009}. 
We highlight some of these aspects in our work and show that we can easily distinguish between subjects' identity from raw gait sensor data. 

\subsection{Manual Feature Extraction}
Much of human motion analysis literature focuses on analysing manually extracted features with statistical analysis or use standard machine learning models such as Support Vector Machines (SVM) for classification. Since extracted features can be strong indicators of emotion \cite{gross_effort-shape_2012}, feature engineering has been used to find an optimal set of gait characteristics that relate with emotions. 
Studies have identified emotions by computing image moments from gait images \cite{das_approach_2015}, extracting temporal, frequency, and temporal-frequency domain features from wearable accelerometer data \cite{zhang_emotion_2016}, and using similarity indices to compare sets of features extracted based on dimensionality reduction techniques such as Principle Component Analysis (PCA) \cite{venture_recognizing_2014}.
On the other hand, there are also studies that use pose-based gait data to extract angular or velocity features. Srivastava et al. \cite{srivastava_robust_2020} extracted angular features such as head tilt, shoulder, right elbow and right knee, and reported high accuracy with shallow machine learning models.

\subsection{Deep Learning Models}
Deep learning models have become increasingly popular due to their ability to automatically extract optimum features that have the potential to capture better the complexity of the gait data than predefined manually extracted features. Commonly applied deep learning models within this domain are Convolutional Neural Networks (CNNs) and Recurrent Neural Networks (RNNs), such as the Long Short-Term Memory (LSTM) models, which have been used for gait recognition and gender identification \cite{saleh_analysis_2021, shahrum_md_guntor_convolutional_2018, zou_deep_2020}, as well as for the detection of abnormalities in a subject's gait \cite{guo_3-d_2019}.

\subsection{Privacy-Preserved Human Motion Analysis}
When exploiting human biological data, there is always a privacy infringement concern even when the intended processing of the data does not aim to extract subject-specific information. Deep learning models can memorise training data and this information leakage can be exploited to reveal subject biometrics \cite{iwasawa_privacy_2017,jung_review_2020}. For example, Song and Shimatikov \cite{song_overlearning_2020} introduces the concept of overlearning where a model trained with the intent of being a gender binary classifier it also unintentionally learns to recognize subjects based on identity and race, although race is not explicitly defined. Therefore, it has become evident that AI approaches should be designed with privacy and ethical concerns taking into consideration \cite{luger_ethics_2020}. This problem is particularly profound in healthcare applications. 
Therefore, it is important to develop algorithms that can dynamically filter streams of data to eliminate unnecessary information that can compromise privacy. Our approach, is using a modular architecture, with a part of it using identity information to guide the learning process to disentangle affects and biometrics. In this way, a part of the network is focused on affect recognition while identify information is inherently filtered out in a measurable manner. 

Recent work has demonstrated that it is possible to separate the latent representation of emotions from the identity representation of a face in a stack of images \cite{kim_deep_2019, wang_survey_2020, liu_hard_2019, zhu_learning_2021}. Typically this method is used to synthesise new data by sampling the latent space representations. Here we suggest exploiting this deep learning approach to disentangle subtle gait features related to affects from the noisy subject-specific features to generate a latent representations of emotions, reducing the complexity of the dataset and enhancing the subjects' privacy in the process. Disentanglement in representation learning refers to the ability to break down data features in key categories and represent them in their own latent spaces. This method is particularly successful in computer vision applications.

Typically to achieve this sort of subject-data disentanglement, a cross-subject training approach is required, where latent representations of the subject data and the target disentanglement data are learned and transferred between each other for different subjects. This type of training has been shown to generalise well and it has been applied in a range of problems, from disentangling identity from human faces \cite{wu_cross-vae_2020} to disentangling pose, expression, and illumination from faces \cite{deng_disentangled_2020}, allowing the model to synthesise a face from different orientations with different expressions and lighting conditions. For human motion applications, this methodology was used to disentangle human pose from images \cite{chen_pman_2021}, to estimate a 3D human mesh from 2D images \cite{sun_human_2019} by learning to disentangle the skeleton from the image, and even for a predictive model of locomotion \cite{mangalam_disentangling_2020}. The major strength of the technique is that it reduces complexity to a low dimension latent representation.

% Our work is closely related to two similar literature pieces that apply an autoencoder-based disentanglement learning method to human motion problems. Aberman et al. \cite{aberman_learning_2019} first applied a similar autoencoder model to learn disentangled representations of human motions, showing that through cross-subject training the motion features can be disentangled from the human skeleton and camera view, and transferred between human subjects by swapping the latent representations of the motions, allowing the motions to be cleanly retargeted without 3D pose estimation. Then Gu et al. \cite{gu_cross-subject_2020} used a similar method to disentangle pathological gait abnormalities from subject data, showing that the reduced complexity of the disentangled latent spaces can improve the classification of the gait patterns over traditional methods. To do this they extend the model with a CNN classifier trained to classify gait patterns from the latent space representations.

\subsection{Explainability in Deep Learning Models}
Deep learning models do not provide direct explanations, and therefore their predictions are difficult to understand and trust, which hinders their adaptation into safety-critical applications. For these reasons, the effort to explain deep learning models has been intensified in the last few years. Explainability methods are categorised into local and global, depending on whether they provide explanations for individual samples or the entire method/group of samples, respectively. Furthermore, methods can be model agnostic or model specific depending on whether they can apply to any model or are tailored to the specific model under investigation \cite{arrieta_explainable_2019}. Gradient-based methods are both local and model specific, since they exploit the neural network activation's mechanisms to directly infer the attributions of the network with relation to a decision. We suggest summing this information across testing samples to obtain global explanations for each class.

\section{Methods}

\begin{figure*}[t]
    \centering
    \subfigure[Model Training Procedure]{
        
        % \centering
        \includegraphics[width=.69\linewidth]{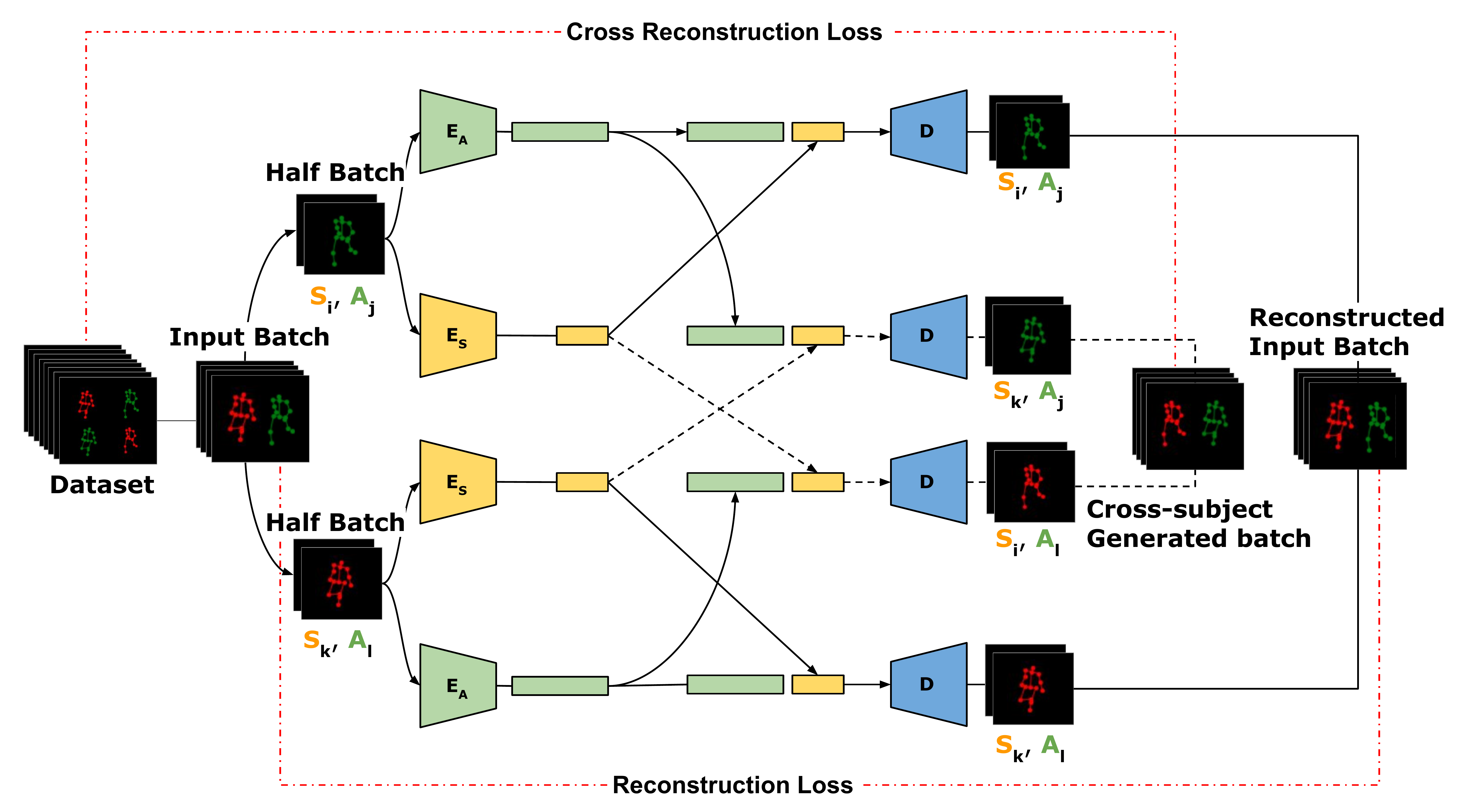}
        
        \label{fig:cross-subject_training}
    }\hspace{-1em}
    \subfigure[Model Architecture]{
        \label{tab:arch}
        \raisebox{35mm}
        \archtab 
    }
    
    \caption{\subref{fig:cross-subject_training} Illustration of cross-subject training procedure. Given two motions $x_{i,j}$ and $x_{k,l}$, the subject and affect features are extracted, their latent spaces are transferred across each other. Then the transferred and original feature combinations are reconstructed, resulting in four different motions, $\hat{x}_{i,j}, \hat{x}_{k,j}, \hat{x}_{i,l}, \hat{x}_{k,l}$. The reconstructions of the original subject/affect pairings $\hat{x}_{i,j}$ and $\hat{x}_{k,l}$ are compared with the original input $x_{i,j}$ and $x_{k,l}$ to calculate the reconstruction loss, and $\hat{x}_{k,j}$ and $\hat{x}_{i,l}$ are compared to the ground truth motions $x_{k,j}$ and $x_{i,l}$ that exist in the full dataset.
    \subref{tab:arch} Architecture summary of Autoencoder model.
    }
    \vspace{-0.9em}
\end{figure*}

\subsection{Cross-subject transfer/Multi-encoder Autoencoder Network}
To perform a disentanglement between subject biometrics and affects we use a separate encoder for each of the features, which transfers the low dimensional feature representations between each other to generate new targets that are a cross over of both features. Figure \ref{fig:cross-subject_training} shows the implemented architecture, which consists of a subject-specific encoder and an affect-specific encoder. A decoder is trained to use the concatenated input of the subject and affect encodings to reconstruct a gait cycle and transfer the emotional state of one person onto another.
In this way, the encoded features are combined during training to generate two different sets of output, a reconstruction batch that matches the original input batch, as well as a cross-subject generated batch that contains new data generated by the autoencoder during training as a result of the cross-subject transfer. Note that for the best results, training using this method requires the ground-truth labels for the cross-subject transfer data to measure the accuracy of the generated data after the latent space transfer.

\subsection{Autoencoder Formulation}
Let $\mathnormal{S}$ and $\mathnormal{A}$ denote the sets of subject biometrics and affect features that form any given gait cycle. Then let  $\left\{x_{i,j}\right\} \in \mathbb{R}^{T \times 3J}$ be a motion, where $\mathnormal{x}$ is a gait cycle described as a subject identity $(\mathnormal{i} \in \mathnormal{S})$, and an affect $(\mathnormal{j} \in \mathnormal{A})$. $\mathnormal{T}$ is the temporal duration of the motion, and $\mathnormal{J}$ is the number of joints each specified by
$\mathnormal{(x,y,z)}$ coordinates.

Our aim is to create an architecture that takes a gait cycle and decomposes it into two factors. Towards this end two encoders $\mathnormal{E_S}$ and $\mathnormal{E_A}$ are trained simultaneously to decompose the gait cycle into two separate latent representation of subject biometrics and affects, respectively. After this decomposition a gait cycle can be defined as $\mathnormal{x_{i,j}} = \mathnormal{D(s_i, a_j)}$ , where $\mathnormal{s_i}$ are subject biometric features encoded by $\mathnormal{E_S} \mapsto{s_i}$, and $\mathnormal{a_j}$ are affect features encoded by $\mathnormal{E_A} \mapsto{a_j}$. $\mathnormal{D}$ is a decoder trained to generate the reconstruction of a decomposed gait cycle, $\mathnormal{D} : (s_i, a_j) \mapsto{\mathnormal{\hat{x}_{i,j}}}$. 

Given two data samples, $x_{i,j}, x_{k,l} \in \mathbb{R}^{T \times3J}$, where $i,k \in S$ and $j,l \in A$, we want to optimise our network such that:
\hspace{-1em}
\begin{equation}
\begin{split}
\forall_{i,k\in S | j,l \in A} \quad \left\|D\left(E_{S}\left(\mathbf{x}_{k, l}\right), E_{A}\left(\mathbf{x}_{i, j}\right)\right) - x_{k,j}\right\| \approx 0
\end{split}
\end{equation}

% This formulation also encourages the encoders $E_S$ and $E_A$ to map inputs of the same class into tight clusters in the corresponding latent maps since we do not make any distinctions between the individual samples within each class.

\subsection{Loss functions}
We use three different types of loss functions to train the autoencoder: a Reconstruction loss, a Cross Reconstruction loss, and a Triplet loss. With a combination of all of them, we can ensure that the reconstructed output is a faithful reconstruction of the original motion while emphasising clustering in the low dimensional encodings.

\subsection{Reconstruction loss}

Firstly, we expect the reconstruction produced by the autoencoder to be faithful to the original input. As such for a sample data point from the training set $\mathcal{X}$, we aim to minimise the difference between the mapped reconstruction of the data sample, and itself using:

\begin{equation}
\mathcal{L}_{\mathrm{rec}}=\mathbb{E}_{\mathrm{x}_{i, j} \sim \mathcal{X}}\left[\left\|D\left(E_{A}\left(\mathbf{x}_{i, j}\right), E_{S}\left(\mathbf{x}_{i, j}\right)\right)-\mathbf{x}_{i, j}\right\|^{2}\right]
\end{equation}

where the full loss for any given batch is calculated using the mean squared error (MSE).

\subsection{Cross-reconstruction loss}

To really strengthen the disentanglement between the different features we also use a cross-reconstruction loss during training. Where given two data samples from the data set $\mathcal{X}$, we transfer the disentangled affect between them to produce two new data samples, computing the reconstruction loss for each using the following:

\begin{equation}
\begin{aligned}
\mathcal{L}_{\text {cross }} &=\mathbb{E}_{\mathbf{x}_{i, j}, \mathbf{x}_{k, l} \sim \mathcal{X} \times \mathcal{X}}\left[\left\|D\left(E_{A}\left(\mathbf{x}_{i, j}\right), E_{S}\left(\mathbf{x}_{k, l}\right)\right)-\mathbf{x}_{i, l}\right\|^{2}\right] \\
&+\mathbb{E}_{\mathbf{x}_{i, j}, \mathbf{x}_{k, l} \sim \mathcal{X} \times \mathcal{X}}\left[\left\|D\left(E_{A}\left(\mathbf{x}_{k, l}\right), E_{S}\left(\mathbf{x}_{i, j}\right)\right)-\mathbf{x}_{k, j}\right\|^{2}\right].
\end{aligned}
\end{equation}
\hspace{1em}

This is possible because the dataset $\mathcal{X}$ contains the ground truth for each affect/subject label combination. Similarly to the reconstruction loss, the loss for a full batch is calculated using MSE.

\subsection{Triplet loss}
Finally, we use a triplet loss function to encourage the autoencoder to cluster the low dimensional encodings tightly. While the reconstruction loss disentangles the subject and affects biometrics, there are no explicit requirements for separating different features in each subject/affect encodings. A triplet loss enforces the separation between the different feature classes in each latent space while also encouraging the same class's features to cluster closer together. 

The triplet loss function is defined with regards to three different data points, an anchor, a positive and a negative. The anchor is the current data point that we are looking at, the positive is another data point of the same feature to whom we want to maximise the similarity, and the negative is a sample of a different feature, for which we want to maximise the difference. Resulting in the following two loss functions for subject and affect:

%\begin{equation}
%\begin{array}{l}
\begin{align}
\mathcal{L}_{\text {trip }}^{S}=\mathbb{E}_{\mathbf{x}_{i, l}, \mathbf{x}_{i, j}, \mathbf{x}_{k, l} \sim \mathcal{X}}[\left\|E_{S}\left(\mathbf{x}_{i, l}\right)-E_{S}\left(\mathbf{x}_{i, j}\right)\right\|-\nonumber\\
\left\|E_{S}\left(\mathbf{x}_{i, l}\right)-E_{S}\left(\mathbf{x}_{k, l}\right)\right\|+\alpha]  \nonumber \\
\nonumber\\
\mathcal{L}_{\text {trip }}^{A}=\mathbb{E}_{\mathbf{x}_{i, l}, \mathbf{x}_{j, l}, \mathbf{x}_{i, k} \sim \mathcal{X}}[\left\|E_{A}\left(\mathbf{x}_{i, l}\right)-E_{A}\left(\mathbf{x}_{j, l}\right)\right\|- \nonumber\\
\left\|E_{A}\left(\mathbf{x}_{i,l}\right)-E_{A}\left(\mathbf{x}_{i, k}\right)\right\|+\alpha] . 
\nonumber\\
\end{align}
%\end{array}
%\end{equation}

Where $\mathbf{x}_{i, l}$ is the anchor, $\mathbf{x}_{i, j}$ and $\mathbf{x}_{j, l}$ are the subject and affect positives, and $\mathbf{x}_{k, l}$ and $\mathbf{x}_{i, k}$ are the subject and affect negatives. $\alpha$ is called a margin that controls the distance between clusters. In other words, it makes the positive data points cluster tighter and separate themselves from the negative data points. Finally, we achieve a significantly faster convergence rate by combining the batch normalisation technique with dropout into a single Independent-Component layer as Chen et al. has suggested to whiten the inputs of a neural network \cite{chen_rethinking_2019}. The integration of which can be seen in Figure/Table \ref{tab:arch} which details the architecture of our model.

\begin{figure}[h]
    \centering
    \includegraphics[width=.95\linewidth]{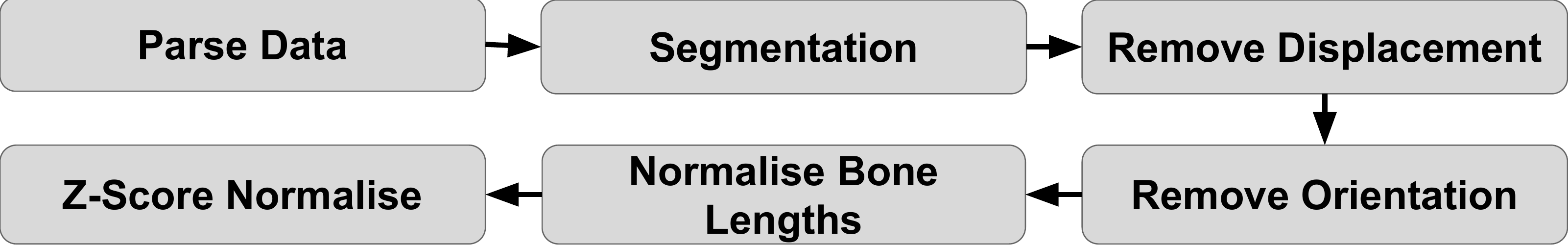}
    \caption{Overview of the preprocessing pipeline.}
    \label{fig:processing_stack}
    \vspace{-1.em}
\end{figure}

\begin{figure*}[htbp]
    % \centering%\quad%\hspace{-1em}
    \begin{center}
    \subcapcentertrue
    \subfigure[$E_A$ validation data latent space.]
    {\includegraphics[width=0.2\textwidth]{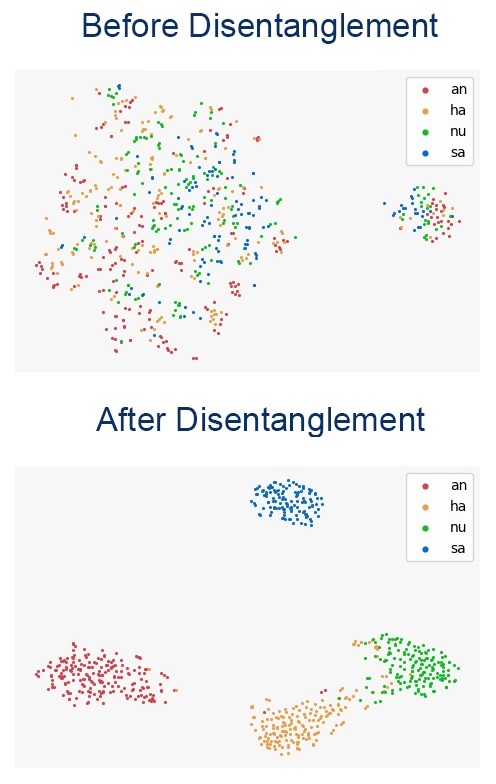}\label{fig:disentanglement}} \quad\hspace{-2em}
    ~
    \subfigure[CNN confusion \newline matrices]
    {\includegraphics[width=0.16\textwidth]{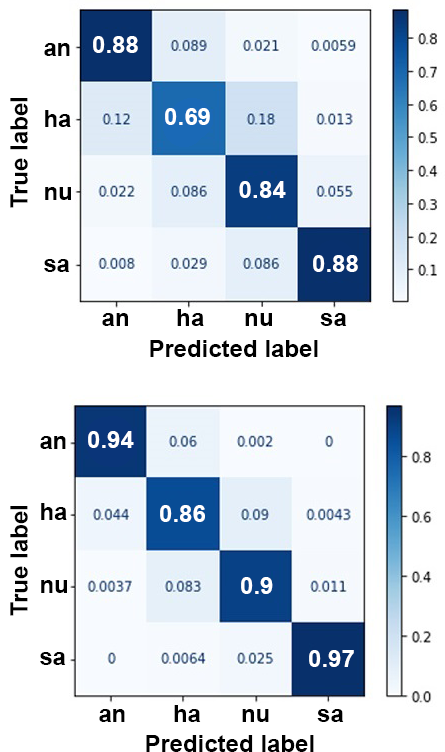}\label{fig:kfold_confusions}} \quad\hspace{-2em}
    ~
    \subfigure[Subject classification.]
    {\raisebox{4mm}{\includegraphics[width=0.30\textwidth]{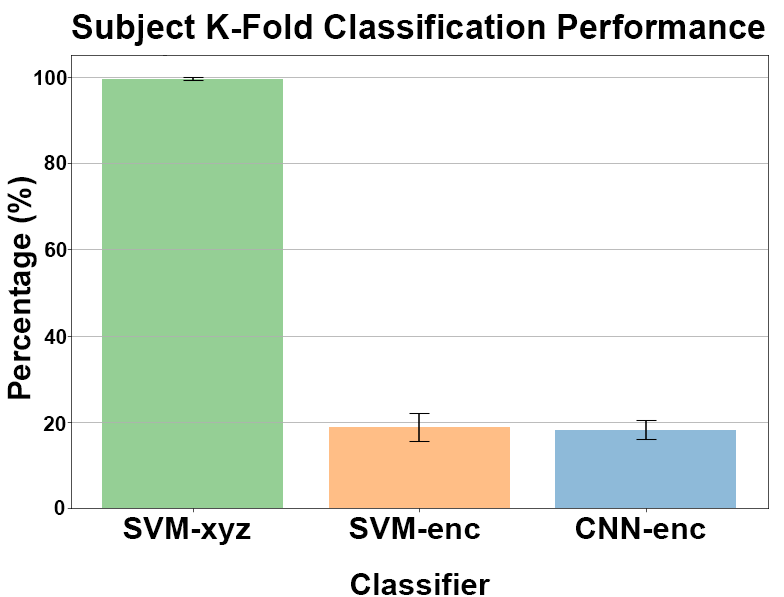}}
    \label{fig:bar_subject}
    }\quad\hspace{-1.5em}
    \subfigure[Affect classification accuracy of each evaluated model.]
    {\raisebox{4mm}{\includegraphics[width=0.30\textwidth]{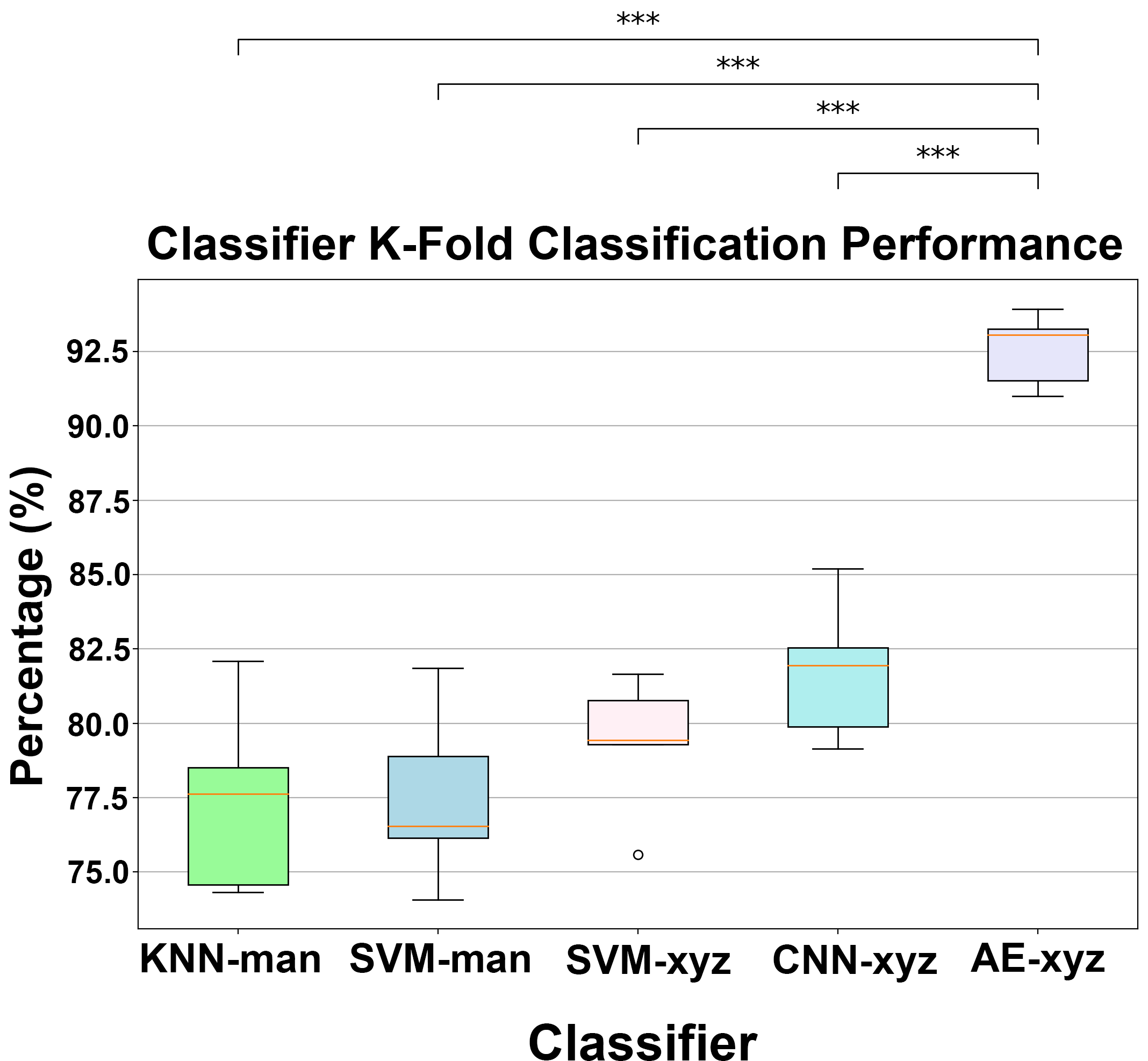}}\label{fig:boxplot_kfold}}\quad%\hspace{-1em}
    ~
    \caption{\subref{fig:disentanglement} A T-SNE visualisation of the $E_A$ latent space for the validation data, before training on the top and after on the bottom, \subref{fig:kfold_confusions} Confusion matrices of the CNN-xyz classifier, with labels an, ha, nu, sa, corresponding to angry, happy, neutral, sad, respectively. The top model trained and validated on the raw xyz data, and the bottom the AE-xyz classifier where the raw xyz validation data is first disentangled by the $E_A$ encoder, and then classified using a CNN classifier trained on the $E_A$ encodings of the train data. \subref{fig:bar_subject} Classification accuracy of subjects using a baseline \textbf{SVM-xyz} that is able to identify subjects with almost 100\% accuracy, compared to models \textbf{SVM-enc} and \textbf{CNN-enc}, trained on the affect latent space rather than raw data. \subref{fig:boxplot_kfold} Classification accuracy of each model represented as boxplots, plotted from the classification accuracy of each fold in the Stratified K-Fold cross-validation experiment, with a pairwise statistical comparison of each model with the AE-xyz model indicated above the figure.}
    \vspace{-1.1em}
    \end{center}
\end{figure*}

\subsection{Guided Grad-Cam for Model Explainability}
The ability to explain the model decisions and link them to the gait cycle is a key factor in enhancing confidence in the model and adopting it in healthcare applications. Here we propose to exploit a local interpretability method to examine global properties of the model.  
To evaluate how our model attributes each of the joint signals to determine the final classification we make use of Guided Grad-Cam.
Grad-CAM is a convolutional network visualisation method that exploits the fact that convolutional layers retain spatial information, and layers deeper within a network aim to capture high-frequency details. As such, the final convolutional layer of a network is expected to be a good compromise between high-level semantic and detailed spatial information. Grad-CAM computes the gradients for the output of a class label with respect to the feature map of a convolutional layer. These gradients are then global average pooled and passed through a ReLU function to reduce the impact of unwanted activations from other classes. Resulting in a low dimension heatmap of the general areas of activation of the same shape as the final convolutional layer. However, our main architecture is an autoencoder and thus the shape of the last convolutional layer is much smaller than the original input.

On the other hand, guided backpropagation visualises individual pixels detected by neurons in the network by backpropagating through ReLU layers and allows for very fine-grain high-resolution visualisation of the activation levels at each timeframe. However, these visualisations are not class-dependent, therefore to combine the best aspects of both approaches, Guided Grad-CAM performs an element-wise multiplication of the high-resolution non-class-dependent visualisation with the low-resolution class-dependent activation heatmap to generate a high-resolution class-dependent activation map that has the same shape as the original input. Retaining the original input shape is crucial for us because, in our case, we cannot feasibly interpolate from a very low-resolution activation map to 45 independent highly variable time series while retaining enough detail to detect changes in activations at specific gait cycle events. With Guided Grad-CAM, on the other hand, we can see changes in neuron activations at a frame-level accuracy allowing us to visualise what happens at different stages of a gait cycle.

\subsection{From Local to Global Explanations}
We are interested in evaluating interpretability across class samples in order to understand global properties of the prediction model. We first split the test data by class so that each sample can be attributed towards its respective class, which in turn gives us an attribution map of how strongly each joint coordinate contributed towards the classification throughout the gait cycle.
To determine a ranking of joint contributions we min-max normalise the absolute attribution values per data sample such that we can determine relative joint importance between all the test gait cycles. Since each joint is made up of three temporal signals that correspond to x, y, and z coordinates of each of the joint. We use the mean of all the signals to represent the overall activation for each of the 15 joints, giving us a single value for each joint in each gait cycle corresponding to the mean activations of that joint throughout. Then by summing the joint activations between gait cycles and dividing by the total number of activations of all joints we can also measure a percentage of how much each joint contributes towards the total activation.

\section{Evaluation Methodology}
We evaluated our proposed model, referred to as \textbf{AE-xyz} against a CNN model, referred to as \textbf{CNN-xyz}, on the raw joint time-series data, and three shallow baseline models that have been used in affect recognition based on human motion data \cite{srivastava_robust_2020}. These are a K Nearest Neighbours (KNN) classifier and a Support Vector Machine classifier trained on the manually engineered angular and velocity features, referred to as \textbf{KNN-man} and \textbf{SVM-man}, as well as an \textbf{SVM-xyz} model that was trained on the raw data. As a baseline the set of raw features extracted were the mean, std, min and max of the key angles such as knees or elbows, as well as velocities of key joints in the arms and legs, all of which have been shown to be useful features for affect recognition \cite{srivastava_robust_2020}. 

We performed an intra-subject stratified K-fold cross validation, where the entire dataset was split evenly into 5 different folds, ensuring an even split of subject/affect pairs between each of the folds. This kind of evaluation gives us an insight into how capable the model is at generalising to unseen gait cycles generated by known subjects.

\subsection{Datasets}
To test our classifier models' capabilities, we are making use of a publicly available motion capture library \cite{ma_motion_2006}. The dataset contains a set of motion captured movements of 30 non-professional subjects (15 male, 15 female, mean age 22, ranging from 17 to 29 years) performing various motions such as walking or throwing, labelled based on their identity, gender and emotion. The subjects acted out scripts specially prepared to stimulate target emotional responses. Our work's focal area will be the dataset's subject and emotion labels, specifically relating to gait motions, with a mean and standard deviation of $107 \pm 15.5$ gait cycle samples, respectively, per subject. The subjects acted out scripts specially prepared to stimulate target emotional responses. The motions were captured using retroreflective markers and a state-of-the-art motion capture system. The raw motion capture data was post-processed through a 3D animation software where the key joint positions from the motion data were projected to outline the human skeleton.

\subsection{Preprocessing}
The preprocessing pipeline is summarised in Figure \ref{fig:processing_stack}. Firstly, it requires the segmentation of the data into gait cycles, defined as two consecutive heel strikes of the same foot. We extract the gait cycles by detecting points of minima in the acceleration of the right ankle.  We use Single Spectrum Analysis (SSA) \cite{golyandina_singular_2018,gu_markerless_2018}, to smooth the acceleration signal for consistent and clean peak detection. To normalise the gait cycles four more steps are required: a) The removal of the global displacement and rotation, to ensure consistency in the coordinate space between the data samples. b) The bone lengths between the subjects are normalised such that the model is encouraged to learn the difference in subject gait styles rather than differences in height, and c) a z-score normalisation is performed such that all of the gait cycles have 0 mean and 1 standard deviation. 

\subsection{Displacement Removal}
To remove the global displacement of the subject during the gait cycle we take the position of a root joint, in our case the center hip joint, and subtract its position from all the other joints for each frame of the gait cycle. Let $x \in \mathbb{R}^{T \times 3J}$ be a motion, where $\mathnormal{x}$ is a gait cycle, $\mathnormal{T}$ is the temporal duration of the motion, and $\mathnormal{J}$ is the number of joints each specified by $\mathnormal{(x\hat{i},y\hat{j},z\hat{k})}$ coordinates. Then $x^{jt}$, is a joint $(\mathnormal{j} \in \mathnormal{J})$ at time frame $(\mathnormal{t} \in \mathnormal{T})$, of the gait cycle. Given a root joint ($\mathnormal{r} \in \mathnormal{J}$), then the new position $\dot{x} ^{jt}$ of the joint without global displacement is given by:

%\[J = J - {j_r}\]
% \[j{'_i} = j{}_i - {j_r}\]
% \[x{'_{jt}} = {x^{jt}} - {x^{rt}}\]
\begin{equation}
    {\dot{x} ^{jt}} = {x^{jt}} - {x^{rt}}
\end{equation}

To remove rotation effects from motion $\dot{x}$, we need to project the coordinates of each joint to a new set of axes, with the facing direction of the skeleton as one of the axes. To extract the forward facing vector we exploit the fact that in general the body's facing is perpendicular to the hip, and as such we define a limb vector ${\boldsymbol{l}_{hp}} \in L$, as a valid connection between two joints, representing the skeleton's bone, where $L$ is the set of all valid joint connections, and $\left\{h,p\right\} \in J$ representing the left and right hip joints respectively. We can derive ${\boldsymbol{l}_{hp}}$ from motion $\dot{x}$ at any given point in time $t$ using:

% \[{l_{(hip)}} = \frac{{{j_{(r\_hip)}} - {j_{(l\_hip)}}}}{{\left\| {{j_{(r\_hip)}} - {j_{(l\_hip)}}} \right\|}}\]

\begin{equation}
    {\boldsymbol{l}_{hp}} = \frac{{{x^{ht}} - {x^{pt}}}}{{\left\| {{x^{ht}} - {x^{pt}}} \right\|}}
\end{equation}

From which we can find the direction vector $\boldsymbol{d}$ by finding the vector perpendicular to $\boldsymbol{l}_{hp}$ and $\hat{\boldsymbol{j}}$, a unit vector representing the up direction.

\begin{equation}
    \boldsymbol{d} = {\boldsymbol{l}_{hp}} \times \hat{\boldsymbol{j}}
\end{equation}

Using all this we can define a change of basis matrix $\mathnormal{C}$, which will transform the joint positions to a new set of axes, with the skeleton facing the y-axis, effectively removing the effects of the z-axis rotation of the motion. We define $\mathnormal{C}$ as:

\begin{equation}
    C = \left[ \begin{array}{l}
    \boldsymbol{l}_{hp}\\
    \boldsymbol{d}\\
    {\boldsymbol{\hat j}}
    \end{array} \right],
\end{equation}

% \[{C^{ - 1}}J = {J'}\]
% \[{C^{ - 1}}{j_i} = j{'_i}\]

and then using it we can define a new position for a joint, $\ddot{x}^{jt}$, where the effects of global displacement and rotation are removed, by solving the following linear equation:

\begin{equation}
    {C^{ - 1}}{\dot{x}^{jt}} = \ddot{x}^{jt}
\end{equation}

\subsection{Bone Normalisation}
To normalize the skeleton bone lengths between the subjects, we define each bone of the skeleton as limb vectors $\boldsymbol{l}_{pc} = \ddot{x}^{ct} - \ddot{x}^{pt}$, where $p \in J$ is a parent joint and $c \in J$ the child joint. For the motions to remain unchanged it is crucial that the joints are rescaled in a hierarchical order starting from the base joint, i.e. to rescale the arm, start with the shoulder, then upper arm and finally the forearm. Each limb vector is rescaled to the mean length of the limb within the dataset $\bar{\boldsymbol{l}}_{pc}$, which can then be used to calculate a scaling ratio $\alpha$ for each limb:

\begin{equation}
\alpha = \frac{{{\bar{\boldsymbol{l}}_{pc}}}}{{\left\| {{\boldsymbol{l}_{pc}}} \right\|}}
\end{equation}

Using this scaling ratio a joint can then be rescaled to the new position $\acute{x}^{jt}$ using:

\begin{equation}\acute{x}^{jt} = \alpha \cdot {\boldsymbol{l}_{pc}} + \ddot{x}^{pt},\end{equation}

where it is now scale, rotation and displacement independent.

\section{Results}
This section aims to evaluate the ability of the proposed method to disentangle biometrics and affects as well as to quantify the ability of the algorithm to preserve the privacy of the subjects and explain the network decisions with relation to the gait cycle and the joints contributions. 

\begin{table}[htbp]
\centering
\caption{Accuracy - Quantitative Results (\%) for Stratified K-Fold Cross Validation}\label{tab:kfold-acc}
\setlength\tabcolsep{5pt} % default value: 6pt
\begin{tabular}{ccccccccclllllll}
\cline{1-9}
\multirow{2}{*}{Model} & \multicolumn{2}{c}{Angry}                   & \multicolumn{2}{c}{Happy}                   & \multicolumn{2}{c}{Neutral}                 & \multicolumn{2}{c}{Sad}\\ \cline{2-9}
                       & Acc                  & Std                  & Acc                  & Std                  & Acc                  & Std                  & Acc                  & Std\\ \cline{1-9}
KNN-man                & 79.71                & 3.58                 & 68.42                & 5.52                 & 75.41                & 4.63                 & 88.17                & 2.35\\
SVM-man                & 80.66                & 3.61                 & 68.89                & 5.59                 & 74.96                & 6.38                 & 87.1                 & 1.83\\
SVM-xyz                & 87.03                & 3.48                 & 69.62                & 3.36                 & 76.2                 & 3.36                 & 85.19                & 4.16\\
CNN-xyz                & 88.4                 & 6.11                 & 68.77                & 9.24                 & 83.62                & 3.97                 & 87.74                & 3.55\\
AE-xyz                 & \textbf{93.82}       & 2.34                 & \textbf{86.17}       & 3.99                 & \textbf{90.25}       & 2.89                 & \textbf{96.82}       & 2.06\\ \cline{1-9}
\end{tabular}
% \vspace{-15mm}
\end{table}

\begin{table}[htbp]
\centering
\caption{F1 Scores - Quantitative Results (\%) for Stratified K-Fold Cross Validation}\label{tab:kfold-f1}
\setlength\tabcolsep{3.5pt} % default value: 6pt
\begin{tabular}{ccccccccccccclllllll}
\cline{1-13}
\multirow{2}{*}{Model} & \multicolumn{3}{c}{Angry}                                          & \multicolumn{3}{c}{Happy}                                          & \multicolumn{3}{c}{Neutral}                                        & \multicolumn{3}{c}{Sad}                                        \\ \cline{2-13}
                       & Pre                  & Rec                  & F1                   & Pre                  & Re                   & F1                   & Pre                  & Rec                  & F1                   & Pre                  & Rec                  & F1               \\ \cline{1-13}
KNN-man                & 82                   & 80                   & 81                   & 71                   & 68                   & 70                   & 72                   & 75                   & 74                   & 86                   & 88                   & 87               \\
SVM-man                & 80                   & 81                   & 80                   & 71                   & 69                   & 70                   & 72                   & 75                   & 74                   & 88                   & 87                   & 88               \\
SVM-xyz                & 83                   & 87                   & 85                   & 72                   & 70                   & 71                   & 76                   & 76                   & 76                   & 88                   & 85                   & 87               \\
CNN-xyz                & 87                   & 88                   & 88                   & 78                   & 69                   & 73                   & 74                   & 84                   & 78                   & 90                   & 88                   & 89               \\
AE-xyz                 & \textbf{95}          & \textbf{93}          & \textbf{94}          & \textbf{88}          & \textbf{90}          & \textbf{89}          & \textbf{92}          & \textbf{91}          & \textbf{91}          & \textbf{96}          & \textbf{97}          & \textbf{96}      \\ \cline{1-13}
\end{tabular}
\end{table}

\begin{figure*}[h]
\centering
\subfigure[Joint Contributions]{
        
        % \centering
        \includegraphics[width=.65 \linewidth]{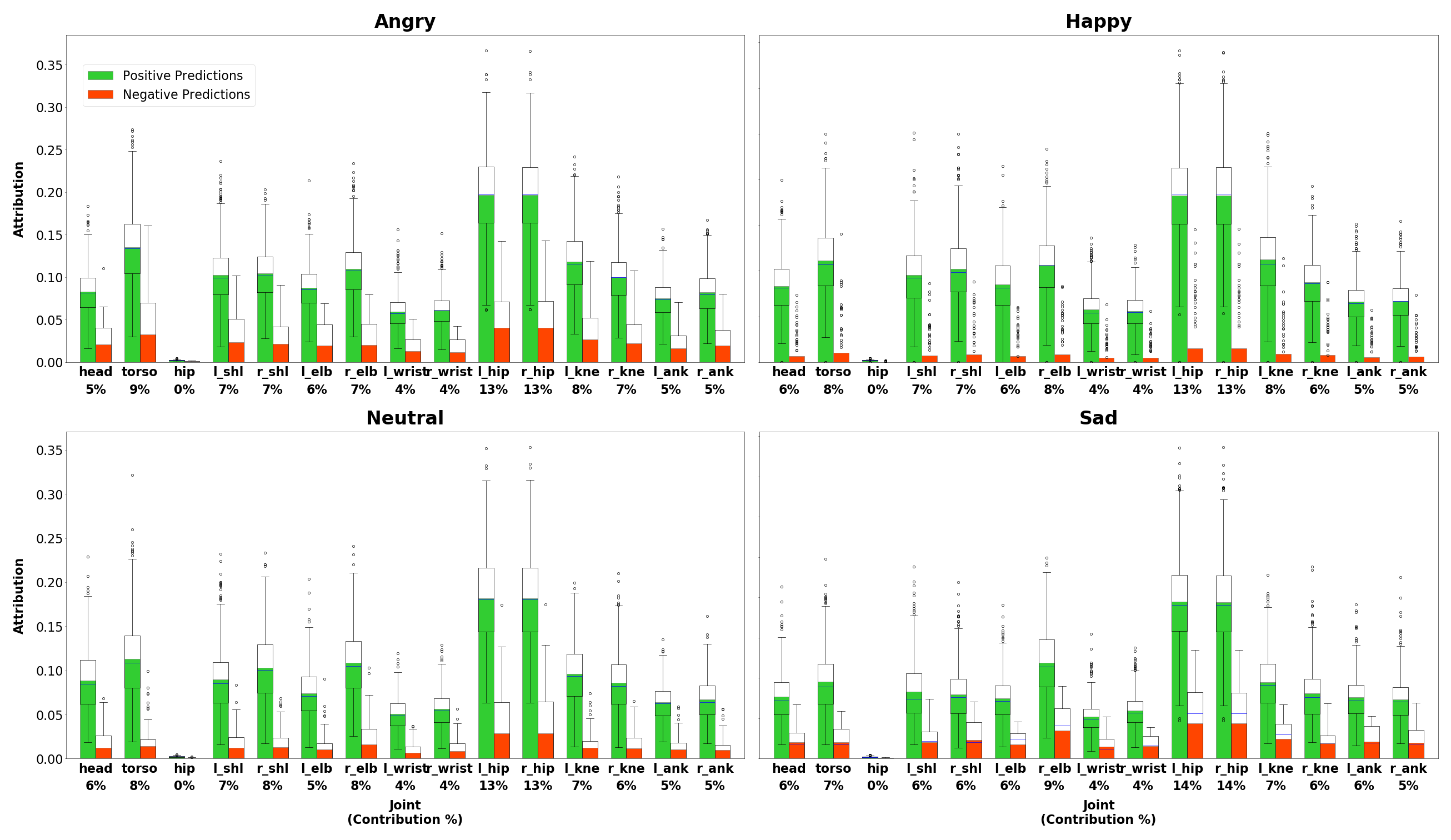}
        
        \label{fig:joint_contribution}
    }
    ~\hspace{-1.7em}
\subfigure[Gait Cycle Activations]{
        
        % \centering
        \raisebox{0mm}{\includegraphics[width=.33\linewidth]{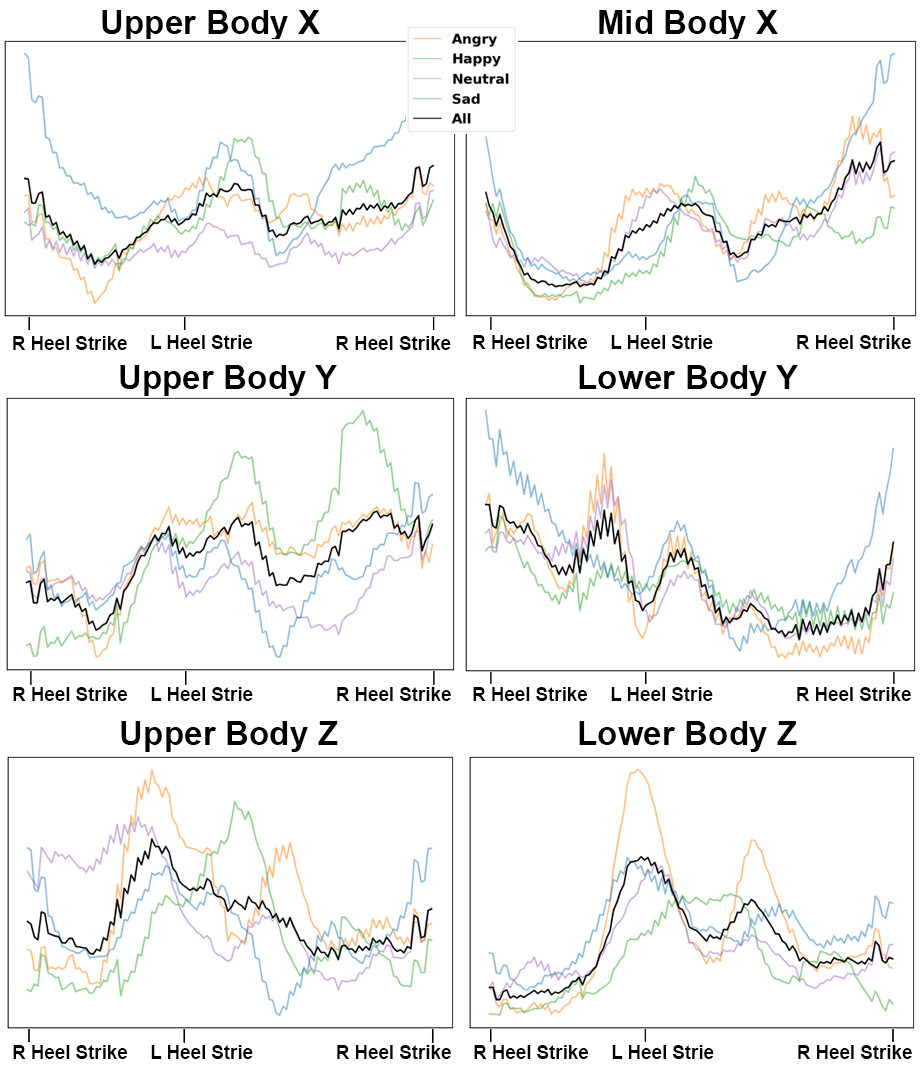}}

        \label{fig:cycle_activations}
    }

\label{fig:interpretability}
\caption{\subref{fig:joint_contribution} A bar chart with the mean total attributions of each joint between the test samples for each emotion and a box plot overlaid to show the distributions of joint attributions between subjects. Prefixes r\_ and l\_ represent the right and left side of the subject's body and joint abbreviations shl, elb, wrist, hip, kne, ank correspond to shoulders, elbow, wrist, hip, knee and ankle respecively. \subref{fig:cycle_activations} A visualisation of the attributions of joint groups, for each emotion individually as well as the mean, over a gait cycle consisting of 128 frames. The upper body consists of the head, neck, shoulders, elbows and wrists. The mid body comprises the hips and the lower body of the knees and ankles. Only the axes with the majority of joint movements, and as a result, the majority of attributions, are presented.}

\end{figure*}

\subsection{Affect Classification}

The proposed methodology improved the mean classification accuracy more than 7\% and the difference in performance between the \textbf{CNN-xyz} and \textbf{AE-xyz} was statistically significant with $(P \leq 0.005)$, achieving a 92.6\% mean accuracy and 93 mean F1 score over the CNN's 84.6\% mean accuracy and 82 mean F1 score. From the confusion matrix in Figure \ref{fig:kfold_confusions} we can see that the models' weakest affect was happy in both cases, however \textbf{AE-xyz} misclassified it a lot less than \textbf{CNN-xyz}, and it was overall better across the board. This shows that the Autoencoder is an efficient way of extracting low-dimensional representations of the raw human motion data to reduce complexity and improve classification. Table \ref{tab:kfold-acc} provides an overview of the accuracy scores for each of the models, and Table \ref{tab:kfold-f1} provides an overview of the F1 scores.

in Figure \ref{fig:disentanglement}, a closer inspection of the low dimensional representations formed by the cross-subject Autoencoder reveals that both the Subject-Encoder and Affect-Encoder are forming clearly separated clusters. In fact, before training the low dimensional encodings are randomly intertwined, whereas after training the affects become separated out into their individual clusters with a small amount of overlap between the happy and neutral affects, where the model's predictive ability is slightly lower.

\subsection{Enhancing Privacy Preservation}
Finally, to evaluate the effectiveness of our subject de-identification, we devised an experiment where we first test the subject identification accuracy using the raw joint position data and compare it to the identification accuracy using the disentangled representation. Figure \ref{fig:bar_subject} demonstrates that subjects classification based on the encoded data is low, which confirms that the proposed method is effective. 

% \begin{table}[h]
% \centering
% \caption{}
% \setlength\tabcolsep{4.5pt} % default value: 6pt
% \begin{tabular}{cccccccccc}
% \multicolumn{1}{l}{}     & \multicolumn{4}{c}{mean}                       & \multicolumn{1}{l}{}  & \multicolumn{4}{c}{standard deviation}                            \\ \hline
% \multicolumn{1}{|c|}{}   & an   & ha   & nu   & \multicolumn{1}{c|}{sa}   & \multicolumn{1}{c|}{} & an    & ha    & nu    & \multicolumn{1}{c|}{sa}    \\ \hline
% \multicolumn{1}{|c|}{an} & 0.98 & 0.97 & 0.97 & \multicolumn{1}{c|}{0.96} & \multicolumn{1}{c|}{} & 0.012 & 0.055 & 0.156 & \multicolumn{1}{c|}{0.020} \\
% \multicolumn{1}{|c|}{ha} & 0.97 & 0.97 & 0.97 & \multicolumn{1}{c|}{0.96} & \multicolumn{1}{c|}{} & 0.055 & 0.077 & 0.056 & \multicolumn{1}{c|}{0.056} \\
% \multicolumn{1}{|c|}{nu} & 0.97 & 0.97 & 0.97 & \multicolumn{1}{c|}{0.96} & \multicolumn{1}{c|}{} & 0.156 & 0.056 & 0.017 & \multicolumn{1}{c|}{0.057} \\
% \multicolumn{1}{|c|}{sa} & 0.96 & 0.96 & 0.96 & \multicolumn{1}{c|}{0.97} & \multicolumn{1}{c|}{} & 0.020 & 0.056 & 0.057 & \multicolumn{1}{c|}{0.021} \\ \hline
% \end{tabular}
% \label{tab:correlation}
% \end{table}

\subsection{Joints' Attribution across and within gait cycles}

Figure \ref{fig:joint_contribution} shows the mean total attributions of each joint between the test samples for each emotion and a box plot overlaid to show the distribution of joint attributions between subjects. We find that the model generally balances its attention between the upper and lower body, with the left and right sides of the hips standing out the most from the others in terms of importance. The center hip joint has an attribution of 0, as it is used to normalise the displacement of the skeleton and thus it does not move from the starting position. There is a general symmetry between the left- and right-hand sides of the body, which is expected since human posture normally remains symmetric under different affects. We also notice that for the positively classified samples depicted with the green bars, the model demonstrates larger activations, which reflect certainty in the decision compared to samples classified incorrectly, which result in smaller mean attributions (red bars).
Overall higher activations are associated with the hips, torso, elbows and head, whereas end effector joints, such as wrists and ankles, attract less attention. 

Different phases of the gait cycle are related to the synergetic function of specific groups of muscles and tenons and thus investigating activations across the gait cycle is clinically relevant. Towards this aim we firstly segment the body into three parts, the upper body, mid-body, and lower body, as we found those areas to have the most overlap between joints' activation. The mid-body was specifically separated from the lower body as the hips have shown themselves to be a strong indicator of emotion and their activation pattern is slightly different from the upper and lower body. The head, torso, shoulder, elbow and wrist joints have been combined for the upper body as their activation patterns closely overlap. Then for the lower body, the knees and ankles were chosen as not only do they tend to follow a slightly different pattern from the upper body, but they also line up more with gait cycle events such as heel strikes which is what we would expect as those events' definitions are based on the lower body. Since the motions have also been normalised to be rotation-independent, any given joint will generally have the majority of its motions mapped on one or two axes, whereas the projection to the third axes will be minimal. For example, the movements of the lower body are largely constrained to the X and Y-axis as the rotation within the Z-axis has been removed from the motions. This results in the variances within the Z-axis not attributing much towards the final classification of the gait cycle. Figure \ref{fig:cycle_activations} shows the normalised attributions of the body parts (Upper, Mid, Lower) for each emotion and their mean across the gait cycle. For a total of 45 temporal signals of the joints, We present the normalised joint attributions considering axes with significant attributions.

%\subsection{Joint Contributions}

We also notice a strong relation between gait cycle events such as heel strikes in the lower body and attributions in the X and Z axis for the lower body. For example in Figure \ref{fig:cycle_activations} we see a strong dip in the attribution of the lower body’s Y-axis as the left heel strikes the ground. Furthermore, we note a peak in activations on both ends of the heel strikes with the model paying attention to the manner of how the subject places their foot and lifts the next one. We also see a strong peak in the lower body Z-axis as the left heel strikes the ground and the right toe-off happens, with continuous activations during the swing phase of the right foot. With the upper body we observe peaks in activations around the left heel strike events with more consistent activations throughout, but this time rather than peaking at the point of the heel strike the model activates stronger earlier when one swing phase terminates and another starts. Presumably, this implies that the model focuses on areas of terminal arm swing. In more expressive emotions with more vivid upper body movement like happy and angry the model activations' are sronger than for less expressive emotions such as neutral and sad, with sad having especially low attributions within the arm swing phase. With the mid-body, we notice more consistent activations between different emotions with strong activations during the cross-over phase where the gait cycle enters into double support and then terminates from it into a right leg swing, presumably the section of the gait cycle where the most expression of the hip tilt happens.

\section{Conclusions}
To our knowledge, our work is the first to disentangle affects and biometrics successfully by exploiting a multi-encoder, decoder architecture to map human motion data onto a low dimensional space. We show that this method not only enhances affect recognition performance but also preservers subjects privacy. Furthermore, based on a gradient-based explainability method, we examine global properties of the model that allow us to understand how different joints contribute to the model decision across gait cycles. Future work will focus on evaluating the method in larger datasets and developing an end-to-end training strategy that does not require explicit gait segmentation. 

\bibliography{IEEEtrans}
\end{document}